\documentclass[runningheads]{llncs}

\usepackage{eccv}

\usepackage{eccvabbrv}

\usepackage{graphicx}
\usepackage{booktabs}

\usepackage[accsupp]{axessibility}  

\usepackage{hyperref}

\usepackage{orcidlink}

\usepackage[ruled,vlined]{algorithm2e}
\usepackage{makecell}
\usepackage{multirow}
\usepackage[table]{xcolor}
\usepackage{pgfplots}
\usepackage{subcaption}
\usepackage{wrapfig}
\usepackage{caption}

\begin{document}

\title{Remembering by Reconstructing:\\Domain Incremental Learning With\\Test-Time Training on Video Streams} 

\author{Jonathan Swinnen\orcidlink{0009-0000-9337-0091} \and
Tinne Tuytelaars\orcidlink{0000-0003-3307-9723}}

\authorrunning{ }
\titlerunning{ }

\institute{
ESAT, KU Leuven, Leuven, Belgium\\
\email{\{jonathan.swinnen,tinne.tuytelaars\}@kuleuven.be}}

\maketitle

\begin{abstract}
In this work we introduce a novel approach to domain incremental learning, adapting models over time to evolving, non-stationary data. 
In contrast to other works, we do not attempt to avoid catastrophic forgetting, but rather allow it and exploit it. 
Our model combines a main task head with a self-supervised masked autoencoder (MAE) head. We then learn domain-specific LoRA adapters during incremental training. Each adapter specializes to its domain, naturally inducing forgetting on other domains in both heads. At inference, we perform online test-time training on the self-supervised MAE head to identify which LoRAs best matches the current input, so the model can `remember' the domain again. Our scheme is especially well-suited to real-world streaming data, such as video, where consecutive samples are highly correlated and domain shifts are gradual. 
We demonstrate our method on domain-incremental action recognition and semantic segmentation tasks. Our code and data will be published upon publication.

\end{abstract}

\section{Introduction}
\label{sec:intro}

Today, most machine learning models are trained on large, static datasets, and assume that the test-time data will resemble the training distribution. While this often leads to strong performance, models struggle when this assumption is violated. In many practical scenarios, the data at deployment may differ from the training domain, or performance could be improved by adapting or “personalizing” the model to the specific deployment environment. The challenge becomes even greater in evolving environments, where the data distribution changes over time. In such cases, adapting to the target domain once is insufficient. Models must instead be incrementally updated as new domains are encountered,
making sure that performance does not degrade 
on previous domains  (so-called catastrophic forgetting \cite{mccloskey1989catastrophic}).
This is precisely the challenge studied in Continual Learning (CL)~\cite{van2019three, wang2024comprehensive}, and Domain Incremental Learning (DIL) in particular.

In standard DIL setups, the model is trained on a sequence of domains one by one.
At test time, the model is not informed which domain a given test sample belongs to, 
treating them all as IID samples from the joint distribution.
This, however, is often not the case in practice. Real-world deployments often involve continuous data streams, such as video feeds,
with subsequent samples highly correlated over time and domain shifts occurring  gradually. This raises the question: {\em can we exploit this temporal continuity to improve adaptation of an incrementally trained model during test-time domain shifts?}

\begin{figure}
[tb]
    \centering
    \includegraphics[width=0.9\textwidth]{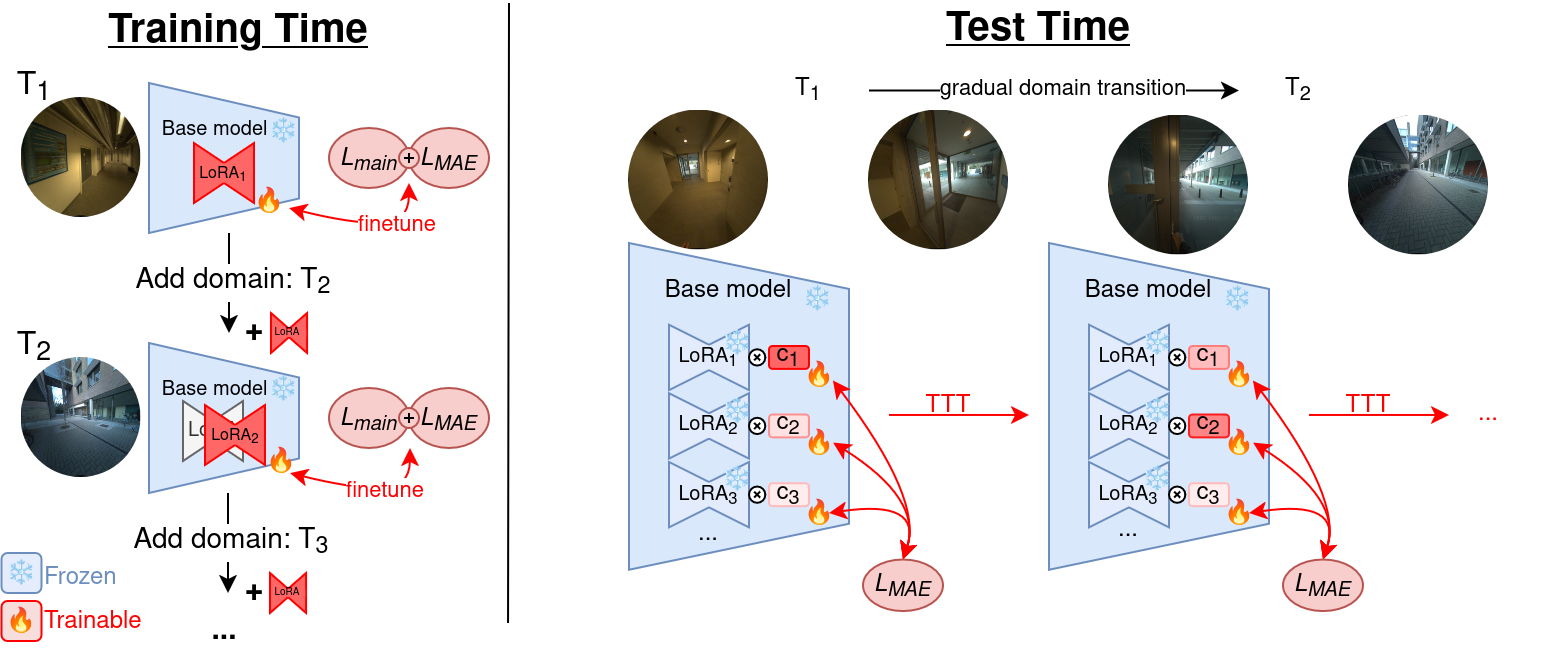}
    \caption{A general overview of our domain incremental learning method: At training-time (left), we train a new LoRA for each new domain on both the main downstream task and an auxiliary MAE task. At test-time (right), we use a weighted combination of all previously trained LoRAs and optimize their weighting coefficients online using test-time training on the MAE task.}
    \label{fig:overview}
\end{figure}

To explore this idea, we were inspired by prior research on test-time training (TTT) \cite{gandelsman2022test, wang2025test} using masked autoencoders (MAEs) \cite{he2022masked, tong2022videomae}. 
They show 
that a model equipped with an auxiliary MAE head can improve classification accuracy on a test sample by first finetuning the MAE on this test sample. Follow-up work \cite{wang2025test} extended this idea to streaming video data, leveraging 'locality': the notion that consecutive frames are highly correlated and that adapting on one frame also benefits nearby ones. However, it was also observed that adapting the model to local frames causes degraded performance when evaluating on frames further away. Rather than viewing this as a limitation, \cite{wang2025test} showed that this forgetting is in fact beneficial and necessary to achieve optimal performance on the current, local frames, challenging the conventional views on continual learning.

In our domain-incremental setting, we 
take this idea one step further. 
The goal is 
to personalize our model to specific domains through incremental training. We accept the fact that an optimal model for the current domain might need to forget domains that are currently not important.
This is fine, as long as we can
quickly adapt back to past domains when we  encounter them again at test time. 

We achieve this as illustrated in \cref{fig:overview}. We use a similar dual-headed architecture composed of a main task head and a self-supervised MAE head as proposed in~\cite{gandelsman2022test, wang2025test}. During incremental training, we sequentially learn a set of domain-specific LoRAs, each fine-tuned to achieve optimal performance within its respective domain. This process naturally introduces forgetting across the other domains as the model specializes. Notably, this forgetting affects not only the main task head but also the MAE head, which we can exploit.

Rather than attempting to explicitly select a single LoRA at test-time \cite{wistuba2023continual} or naively merging them \cite{chitale2023task}, our model maintains a weighted combination of all domain-specific LoRAs. During inference, we perform online test-time training on the MAE head to optimize the weighting coefficients, keeping everything else frozen. In doing so, the MAE reconstruction loss naturally steers the model toward the LoRAs that match the current domain, reversing the forgetting that occurred in the incremental training stage. 

Training only this small set of weighting coefficients is much easier than adapting the full model at test-time. Moreover, because we operate on streaming data where consecutive frames are highly correlated, the weighting coefficients do not need to be updated every single frame. 
This enables our method to adapt rapidly and efficiently
to gradual domain shifts in a test stream.\\
\\
\textbf{In summary}, we propose a domain-incremental learning and online adaptation method leveraging domain specific LoRAs and test-time MAE training to adapt to domain changes during inference on a video stream. We test our method on domain-incremental video action recognition and semantic segmentation tasks.

\section{Related Work}

\label{sec:related}

\subsection{Continual Test-Time Adaptation}

Continual Test-Time Adaptation (CTTA) aims to adapt a model to distribution shifts while performing inference on a data stream. Many approaches maintain an online-updated student model together with an exponential moving average teacher that generates pseudo-labels by averaging predictions from multiple augmented views of the current sample \cite{wang2022continual, tian2024parameter, park2025hybrid}.
Another line of work follows Test-Time Training (TTT), where the model is optimized at test-time using an auxiliary self-supervised objective in addition to the main task. Typical auxiliary tasks include rotation prediction \cite{sun2020test} or masked autoencoding \cite{gandelsman2022test, wang2025test}. During inference, the model first adapts to incoming samples using the self-supervised objective before making predictions.
Most CTTA works study covariate shifts such as synthetic corruptions or changes in lighting and weather conditions. However, the effectiveness of these methods is often highly sensitive to hyperparameters and the specific type of shift, and they can struggle under more complex distribution changes such as label shift or spurious correlation shift \cite{zhao2023pitfalls}.

\subsection{Domain Incremental Learning}
Domain Incremental Learning (DIL) \cite{van2019three} is a type of Continual Learning (CL) \cite{wang2024comprehensive, de2021continual} where a model is incrementally trained on new domains and must maintain high performance on all of them. Unlike CTTA, DIL typically involves offline supervised training, which allows for learning more complex distribution shifts and acquiring additional domain-specific knowledge. The central challenge is catastrophic forgetting \cite{mccloskey1989catastrophic}, where adaptation to new domains overwrites previously acquired knowledge. 
One way of mitigating this effect is through the use of pre-trained models and parameter-efficient finetuning (PEFT) schemes. Rather than modifying all model parameters, PEFT restricts adaptation to small, dedicated parameter subsets, such as prompts \cite{lester2021power} or low-rank adapters \cite{hu2022lora}. By localizing domain-specific updates, interference with previously learned knowledge is naturally reduced. In Task Incremental settings, where the task identity is provided at test-time, the appropriate PEFT modules can be selectively activated or deactivated, allowing the model to retain all previously learned knowledge without forgetting. However, in Domain Incremental scenarios where task identity is unknown at test-time, this is not possible. 
In this case, the knowledge of the separate PEFT modules must either be fused or combined to make the model task-agnostic \cite{chitale2023task, gao2023unified, liang2024inflora}, potentially causing forgetting, or the right PEFT modules should be selected at test-time based on the input \cite{wang2022learning, wang2022dualprompt, wistuba2023continual, wang2025hide}, which can be a challenge.
\section{Method}
\label{sec:method}
\subsection{Model Architecture}
\label{sec:method/arch}

\begin{figure}[htbp!]
    \centering
\includegraphics[width=0.75\textwidth]{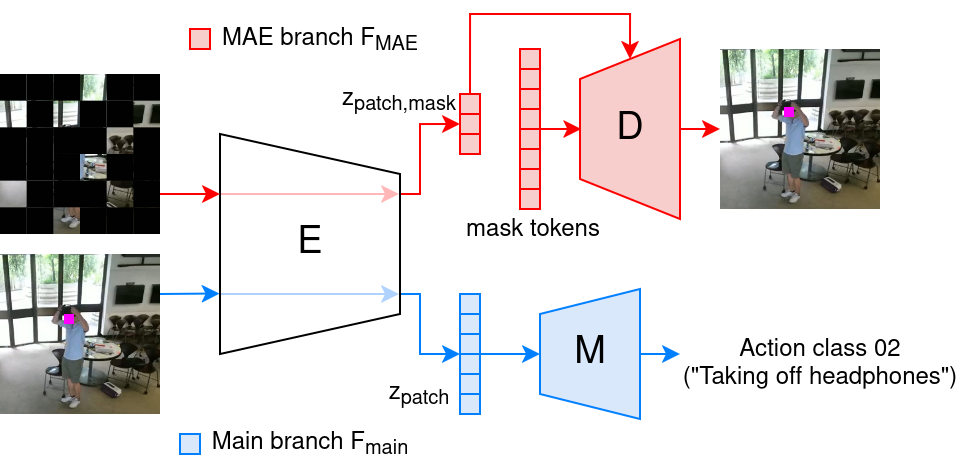}
    \caption{Illustration of our model architecture for the case of video action recognition. The upper branch consists of an encoder E and a decoder D, which are trained as an MAE. The lower branch shares the same encoder but uses a different head M for the main downstream task.}
    \label{fig:arch}
\end{figure}

Similarly to TTT-MAE \cite{gandelsman2022test, wang2025test}, we adopt a model with a shared backbone encoder $E$ followed by two parallel task-specific decoders. The first decoder, $D$, is used for the self-supervised MAE objective \cite{he2022masked}, while the second decoder, $M$, is dedicated to the downstream task (e.g., classification or semantic segmentation). An overview of the architecture is shown in \cref{fig:arch}.

In our experiments, the MAE branch $D \circ E$ follows a transformer-based design, as is common for masked autoencoders. For action recognition, we use a spatiotemporal MAE similar to \cite{tong2022videomae, feichtenhofer2022masked}, where tokens attend both spatially within a frame and temporally across a window of past frames, enabling the encoder to model motion dynamics in addition to spatial structure, which is crucial for recognizing actions. For semantic segmentation, we instead opt for a simpler, more lightweight MAE which processes each frame independently, without temporal attention.

The downstream head $M$ is task-dependent. For action recognition, we use an attentive classifier head, inspired by \cite{chen2024context, bardes2024revisiting}, consisting of four cross-attention blocks which take a learnable classifier token as query and the output features of the encoder as keys and values. For semantic segmentation, we adopt the Mask2Former decoder \cite{Cheng2021MaskedattentionMT}. Further architectural and implementation details are provided in Appendix~\ref{app:arch}.

\subsection{Offline Training}
\subsubsection{Initial task pre-training:}
We start by initializing our encoder $E$ and decoder $D$ parameters $\theta^E$ and $\theta^D$ from a pre-trained masked autoencoder checkpoint, and we randomly initialize the parameters $\theta^M$ of the main task head $M$.
We do not discard the MAE decoder, instead we keep it and continue finetuning it alongside the main task head. 

Each training step, we sample a batch of samples $\mathbf{x}$ and their corresponding labels $\mathbf{y}$. Next, the model first patchifies $\mathbf{x}$ and randomly masks a fraction $r_{mask}$ of its tokens. The remaining patches $\mathbf{x_{\text{mask}}}$ are passed through the MAE branch $F_{\text{MAE}} = D \circ E$, which reconstructs some of the masked patches $\mathbf{\hat{x}_{\text{dec}}}$. These reconstructed patches are trained to match their corresponding ground truth patches $\mathbf{x}$ using the standard MAE reconstruction loss $\mathcal{L_{\text{MAE}}}(\mathbf{\hat{x}_{\text{dec}}},\mathbf{x})$ \cite{he2022masked}.

Next, we pass the samples through the main task branch $F_\text{main} = M \circ E$, which outputs a prediction $\mathbf{\hat{y}}$, which is trained to minimize the main task loss $\mathcal{L_{\text{main}}}(\mathbf{\hat{y}},\mathbf{y})$. The two losses are added together, with a weighting coefficient $\alpha$ for the MAE loss. Our training objective to train the initial model parameters $\theta_0 = \{\theta^E_0, \theta^M_0, \theta^D_0\}$ can thus be expressed as in \cref{eq:loss}

\begin{equation}
\label{eq:loss}
\begin{split}
\theta_0 = \arg\min_\theta \mathbb{E}_{(\mathbf{x},\mathbf{y}) \sim T_0} \big[\mathcal{L_{\text{main}}}(F_{\text{main}}(\mathbf{x}), \mathbf{y}) + \alpha \mathcal{L_{\text{MAE}}}(F_{\text{MAE}}(\mathbf{x_{\text{mask}}}), \mathbf{x})\big]
\end{split}
\end{equation}

\subsubsection{Domain-incremental learning}
After having trained a general model on the first task, we aim to incrementally adapt and specialize the model to new domains.
At any time, when we wish to update our model with knowledge of a new domain task $T_i$, we take our base model parameters $\theta_0$ trained on the initial task and we freeze them. We then initialize new trainable LoRAs $\Delta_i$ for all linear layers in the entire model, and add them to the base model parameters, as in \cref{eq:thetati}.

\begin{equation}
\label{eq:thetati}
    \theta_{i} = \theta_0 + \Delta_i
\end{equation}

We then train the LoRAs on the domain-specific data from $T_i$, using the same training objective as before (\cref{eq:loss}), but now optimizing the LoRAs $\Delta_i$, while keeping $\theta_0$ frozen.

\subsection{Inference and Test-Time-Training}

At test time, the LoRA parameters from all $N$ tasks trained so far are frozen and combined as a weighted sum with weighting coefficients $\mathbf{c}=\{c_1,...,c_N\}$, to get the test-time parameters $\theta_{TT}$ as in \cref{eq:thetattt}.
\begin{align}
\label{eq:thetattt}
    \theta_{TT} &= \theta_0 + \sum_{k=1}^{N}c_k \Delta_k
\end{align}

The question now becomes which weighting coefficients $\mathbf{c}$ we should choose at test time to achieve optimal performance. Selecting one LoRA for a specific domain $T_i$ will yield good performance in this domain, but could induce degraded performance when encountering a different domain $T_j$, which can be seen as some form of 'temporary' forgetting. We can however use this to our advantage, because this forgetting also occurs in the self-supervised MAE head, which can act as a probe revealing when the domain changes and performance degradation occurs. By optimizing the weighting coefficients on the MAE objective using the incoming data stream, as in \cref{eq:ct}, we can 'reverse' this forgetting and adapt the model to the current input $\mathbf{x}_t$. 

\begin{align}
\label{eq:ct}
    \mathbf{c} = \arg\min_\mathbf{c} \mathcal{L_{\text{MAE}}}(F_{\text{MAE}}(\mathbf{x}_{t,\text{mask}}), \mathbf{x}_t)
\end{align}

Only updating this small set of weighting coefficients, is a key advantage of our method, as this converges much faster than re-training the full model. This means we only need to perform a single adaptation step every few frames.

More concretely, once every $n$ time steps, we sample a batch of $m$ recently seen frames from a sliding window buffer $B$ of size $w$, and perform a single update of the weighting coefficients using gradient descent on the MAE loss. 
A smaller $n$ means more updates and potentially faster adaptation, but also increases computational overhead, as more backpropagation steps are performed. A smaller $m$ reduces compute and memory cost for each gradient update, as less frames need to be processed, but could lead to less stable gradients and slower convergence. 

While the coefficients $\mathbf{c}$ are directly optimized using the MAE loss, we first apply a softmax with temperature $\tau$ before using them in the main task branch, as in \cref{eq:thetatttm1,eq:thetatttm2}. This makes the model focus more strongly on the most relevant LoRAs, which could improve performance on the downstream task. 

\begin{align}
\label{eq:thetatttm1}
    \mathbf{\tilde{c}} &= softmax(\mathbf{c/\tau})\\
\label{eq:thetatttm2}
    \theta_{TT,\text{main}} &= \theta_0 + \sum_{k=1}^{N}\tilde{c}_k \Delta_k
\end{align}

Using a small temperature makes the LoRA selection more confident, with $\tau=0$ being equivalent to taking the hardmax, only selecting the LoRA with the highest coefficient. Larger temperatures smooth out the LoRA selection more, with $\tau=\infty$ being equivalent to always taking the average of all LoRAs, regardless of the test-time training. 
We only apply this softmax to the main task coefficients. We do not use it in the MAE branch, as we observed this makes the test-time training convergence more difficult.

\begin{algorithm}
\caption{Online Adaptation with Test-Time training}
\label{alg:online_lora_update}
$\mathbf{c} \gets [1/N, 1/N, ..., 1/N]$ \tcp{initialize model LoRA coeffs to average}
$\text{optimizer.params} \gets \mathbf{c}$ \tcp{only adapt coeffs, rest of model is frozen}
\For{$t = 1$ \KwTo $T$}{
    $x_t \gets $ Next frame from input stream;\\
    \If{$|B|=w$}{
        Remove $x_{t-w}$ from $B$ \tcp{sliding window buffer is full}
    } 
    Save $x_t$ to $B$;\\
    \If{$t \bmod n = 0$}{ \tcp{perform TTT update}
        set $F_{\text{MAE}}$ coeffs $\gets \mathbf{c}$\;
        
        $\mathbf{x_\text{recent}} \gets $ Sample $m$ frames from $B$\;
        
        $\mathbf{x_\text{mask}} \gets $ Randomly mask a fraction $r_{mask}$ of patches in $\mathbf{x_\text{recent}}$\;
        
        $\mathbf{\hat{x}_\text{dec}} \gets F_{\text{MAE}}(\mathbf{x_\text{mask}})$\;
        
        $L \gets \mathcal{L}_{\text{MAE}}(\mathbf{\hat{x}_\text{dec}}, \mathbf{x_\text{recent}})$ \;
        
        $\mathbf{c} \gets \text{optimizer.step}(\mathbf{c}, \nabla_\mathbf{c} L)$         \tcp{update model LoRA coeffs}
        set $F_{\text{main}}$ coeffs $\gets softmax(\mathbf{c}/\tau)$\;
    }

    $\hat{y}_t \gets F_{\text{main}}(x_t)$ \tcp{make prediction}
}
\end{algorithm}

The inference procedure is summarized in \cref{alg:online_lora_update}. We do want to note, that in the case of action recognition, KV-caching \cite{pope2023efficiently} is used to efficiently allow the current frame to attend to previous frames in a streaming setting. Correctly managing this KV-cache requires slight modifications to \cref{alg:online_lora_update}, but the core idea remains the same. The specifics of KV-caching are further discussed in Appendix~\ref{app:kv}. 
\section{Experiments}
\label{sec:experiments}

\subsection{Tasks}
\subsubsection{Video Action Recognition:}
\begin{figure}[t]
    \centering
    \includegraphics[width=\textwidth]{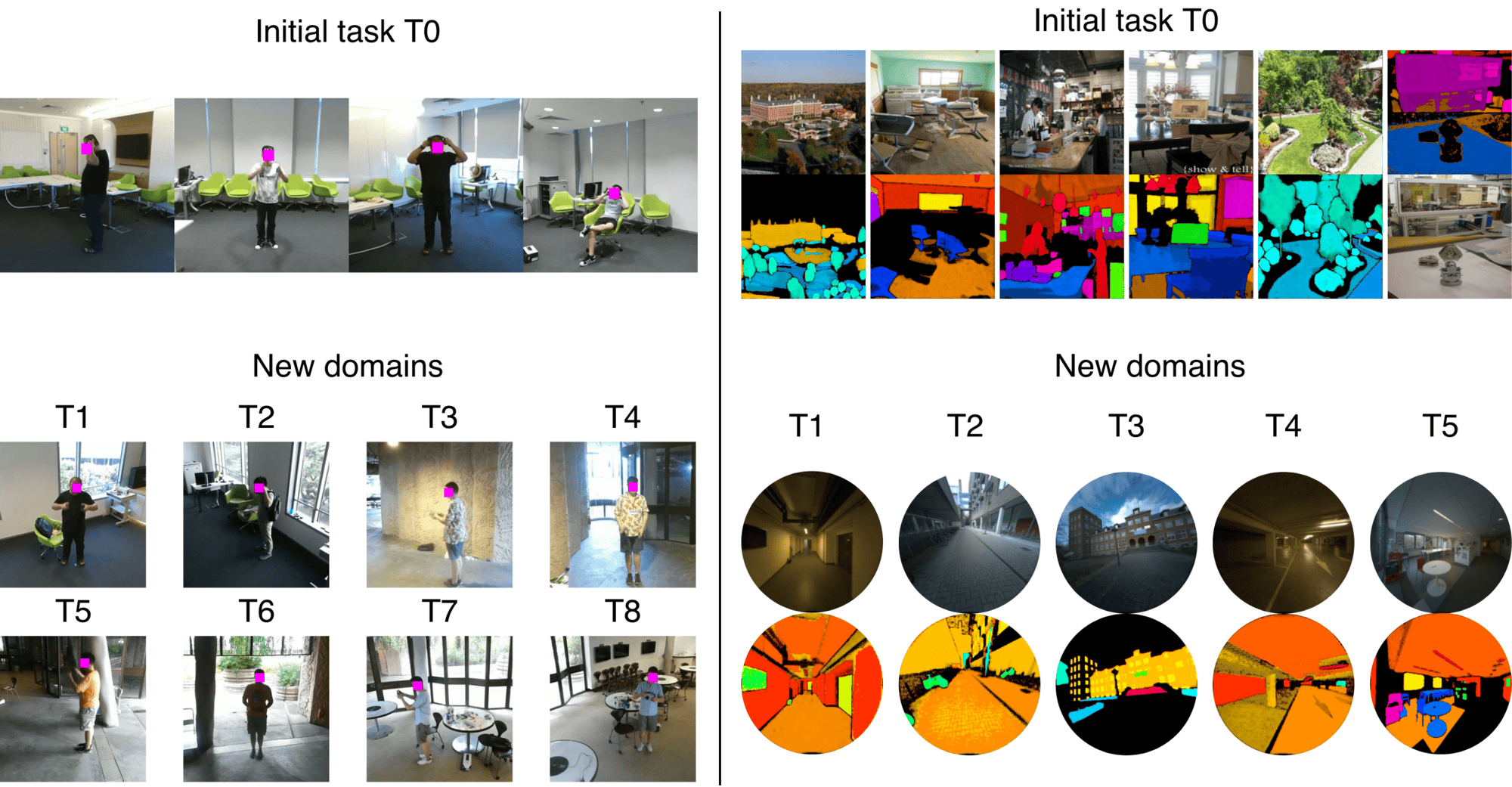}
    \caption{Some example frames illustrating the task sequence of the \textbf{NTU-RGB+D120} dataset (left) and the \textbf{Places365 + Domains-Campus} datasets (right).}
    \label{fig:tasks}
\end{figure}

First, we apply our method to the \textbf{NTU RGB} \textbf{+D120} \cite{shahroudy2016ntu,liu2019ntu} video action recognition task. This dataset consists of short recordings of 120 different classes of actions. For our experiments, we only consider the subset of RGB videos corresponding to the last 60 action classes, which were recorded across multiple environments and camera setups, making them well suited for domain-incremental learning. The initial pre-training task, $T_0$, comprises recordings within a single room. The subsequent tasks, $T_1$ through $T_8$, illustrated in \cref{fig:tasks}, each introduce videos from a novel viewpoint or location, with a fixed camera setup. Specific details on how the task splits are set up are listed in Appendix~\ref{app:ntu}. Since the video clips in this dataset are only a few seconds long, this does not give our test-time training enough time to fully adapt to each domain during evaluation. To address this, we construct longer evaluation sequences by concatenating 60 clips from the same domain. These extended sequences are then shuffled to simulate domain switches

\subsubsection{Semantic Segmentation:}
Secondly, we set up a semantic segmentation experiment. Existing semantic segmentation benchmarks do not simultaneously provide (i) explicit domain splits suitable for continual learning and (ii) a long, continuous test stream that reflects natural, gradual transitions between these domains. Therefore, we collected a new semantic segmentation dataset, \textbf{Domains-Campus}, specifically designed to satisfy these properties. Using Meta ARIA Glasses \cite{engel2023project}, we recorded egocentric training videos while walking around in five distinct environments, each defining a separate domain. In addition, we recorded a longer, continuous test sequence that traverses all training domains in a single walk. This evaluation stream exhibits gradual domain shifts rather than abrupt switches, and also includes transition segments that pass through intermediate, unseen environments which are not in the training set. With the help of SAM3 \cite{carion2025sam}, we extract synthetic segmentation maps for 32 different concepts for each video frame, which we consider as our ground-truth labels for training. The task is thus to mimic the outputs of this teacher model.
The amount of data collected is however not enough on its own to learn the segmentation task properly from scratch. For this reason, we also annotate a subset of roughly 540k images from the \textbf{Places365} dataset \cite{zhou2017places} with the same 32 concepts, which we can use as initial task $T_0$ to pre-train the general base model, after which the model is incrementally specialized to the new domains of the Domains-Campus dataset in the subsequent tasks. \cref{fig:tasks} illustrates some examples of the data for this setup. For more in-depth details about this dataset we also refer to Appendix~\ref{app:campus}.

\subsection{Baselines}

We compare our approach against several baselines. First of all, we report the performance of the intial model after $T_0$ on the task sequence with \textbf{no finetuning} and after \textbf{naïve fine-tuning}. Another baseline is to perform \textbf{LoRA merging} \cite{chitale2023task}, where we simply average the LoRAs. We also include a stronger LoRA-based method, \textbf{CoLoR++} \cite{wistuba2023continual}, which also uses input data to dynamically select the appropriate LoRA at test-time. After training on a domain, its K-Means cluster centers are computed in representation space, and at test time, the nearest cluster to the incoming sample determines the selected LoRA.
Next, we also include \textbf{Replay} \cite{rolnick2019experience}, where each training batch contains an equal mix of samples from the current task and from a replay buffer containing examples from previous tasks. We additionally compare against \textbf{DRIFT} \cite{hu2025video}, which also employs replay and combines it with a knowledge distillation loss to preserve outputs from previous models. Finally, we include a \textbf{task oracle}, which always selects the correct LoRA for the current domain, and \textbf{joint training}, where all tasks are combined and trained on simultaneously.

\subsection{Models}
For the action \textbf{classification task}, we initialize our model using a video masked autoencoder \cite{tong2022videomae} checkpoint pre-trained on the Kinetics400 dataset \cite{kay2017kinetics}. We use a ViT-Tiny model size, using windowed causal attention with window size $w_{DE}=16$, and RoPE positional embeddings \cite{su2024roformer} for the time axis to allow efficient streaming inference. We pre-train this model from scratch using a recipe similar to the one used for VideoMAE \cite{tong2022videomae}. Unlike in the original VideoMAE setup, we apply fully random masking rather than tube masking, as this simpler and more flexible for the test-time training phase. We use an input resolution of $224 \times 224$ at a frame rate of 10 FPS.
Next, we add the main task head, which also uses windowed causal attention with a larger window of 64 frames. We finetune the base model on $T_0$, after which we can start incrementally adding and training LoRAs for each of the 8 domains. During training, we weigh both losses $\mathcal{L_\text{main}}$ and $\mathcal{L_\text{MAE}}$ equally, with $\alpha=1$.
\\
The \textbf{semantic segmentation} experiment uses a model based on Mask2Former \cite{cheng2022masked}. We initialize from a pre-trained Tiny-size checkpoint and we replace its original Swin-Tiny \cite{liu2021swin} backbone with a pre-trained Hiera-Tiny \cite{ryali2023hiera}, which is better suited for masked autoencoding. This model predicts the semantic segmentation map of each frame independently, and again uses a resolution of $224 \times 224$. During training, we scale up the MAE loss by a factor $\alpha = 10$, to not make it negligibly small in comparison to the Mask2Former loss, which is otherwise much larger in magnitude.

We use a LoRA rank of $64$ in all experiments. At inference, unless mentioned otherwise, we perform a TTT update every $n=16$ frames with a learning rate $\eta=0.05$, weight decay $\lambda = 0.03$ and a coefficient temperature of $\tau = 0.2$. For the segmentation model, we use a frame buffer with sliding window size $w=32$, and we randomly sample batches of size $m=16$ with replacement. For the action recognition model, we use a window size $w=16$ and do not use random sampling, instead, we take all $m=16$ consecutive frames in the correct order, which is required for the temporal attention to function correctly. More in-depth technical model architecture implementation details and training recipes are listed in detail in Appendix~\ref{app:train}. 

\subsection{Results}

\Cref{tab:base} reports the final average accuracy of our method compared to the baselines on evaluation streams covering all domains. For reference, the action recognition model trained only on $T_0$ achieves 73.88\% accuracy on its $T_0$ test set, while the semantic segmentation model reaches a macro Dice score of 0.724. Without adaptation, performance drops on new domains due to domain shifts. Naive finetuning offers limited improvement: for segmentation, gains are mostly restricted to the last task, and for action recognition, forgetting can make the average performance worse than the non-finetuned base model.
\\
\begin{table}
    \centering
    \caption{\textbf{Performance of our model compared to other baselines}.
    We show the final average accuracy (dice score for segmentation) and forgetting across $T_1$ to $T_N$. For Domains-Campus, we exclude segments in unseen domains from the scores. We report the mean and standard deviation over 3 random training seeds. For our method, we evaluate each of the training runs with 3 random inference seeds for a total of 9 runs.
    }

    \footnotesize
    \setlength{\tabcolsep}{6pt} 
    \renewcommand{\arraystretch}{1.2}

    \begin{tabular}{c|cc|cc}
    \toprule
        & \multicolumn{2}{c|}{NTU-RGB+D120} & \multicolumn{2}{c}{Domains-Campus} \\
        Method & FAA↑ & F↓ & FAA↑ & F↓ \\
    \midrule
        No FT & $0.645 {\scriptstyle \pm 0.001}$ & $0.000 {\scriptstyle \pm 0.000}$ & $0.302 {\scriptstyle \pm 0.002}$ & $0.000 {\scriptstyle \pm 0.000}$ \\
        Naïve FT & $0.602 {\scriptstyle \pm 0.009}$ & $0.123 {\scriptstyle \pm 0.010}$ & $0.326 {\scriptstyle \pm 0.003}$ & $0.168 {\scriptstyle \pm 0.001}$ \\
    \midrule
        Replay & $0.668 {\scriptstyle \pm 0.007}$ & $0.052 {\scriptstyle \pm 0.005}$ & $0.425 {\scriptstyle \pm 0.006}$ & $0.050 {\scriptstyle \pm 0.009}$ \\
        DRIFT & $0.673 {\scriptstyle \pm 0.004}$ & $0.048 {\scriptstyle \pm 0.004}$ & $0.430 {\scriptstyle \pm 0.004}$ & $0.037 {\scriptstyle \pm 0.009}$ \\
        LoRA merging & $0.669 {\scriptstyle \pm 0.005}$ & $0.050 {\scriptstyle \pm 0.002}$ & $0.410 {\scriptstyle \pm 0.010}$ & $0.029 {\scriptstyle \pm 0.014}$ \\
        CoLoR++ & $0.678 {\scriptstyle \pm 0.010}$ & $\mathbf{0.021 {\scriptstyle \pm 0.006}}$ & $0.448 {\scriptstyle \pm 0.008}$ & $0.023 {\scriptstyle \pm 0.002}$ \\
        Ours & $\mathbf{0.699 {\scriptstyle \pm 0.003}}$ & $0.028 {\scriptstyle \pm 0.003}$ & $\mathbf{0.458 {\scriptstyle \pm 0.007}}$ & $\mathbf{0.011 {\scriptstyle \pm 0.010}}$ \\
    \midrule
        Task oracle & $0.713 {\scriptstyle \pm 0.003}$ & $0.000 {\scriptstyle \pm 0.000}$ & $0.475 {\scriptstyle \pm 0.002}$ & $0.000 {\scriptstyle \pm 0.000}$ \\
        Joint FT & $0.747 {\scriptstyle \pm 0.002}$ & - & $0.450 {\scriptstyle \pm 0.011}$ & -\\
    \bottomrule
    \end{tabular}

    \label{tab:base}
\end{table}
\\
Continual learning approaches mitigate this forgetting, improving overall performance. Our method achieves the strongest results, even surpassing replay-based baselines without storing past data, and comes remarkably close to the task-oracle upper bound. Surprisingly, our method even outperforms joint fine-tuning on the Domains-Campus dataset. We hypothesize that this stems from the expert nature of the individual LoRAs: each one is highly specialized to its own domain and can surpass a single generalist model within their intended setting. This is also suggested by the higher accuracy of the task oracle for this dataset. As a result, a method such as ours, which can consistently use the right LoRAs could also outperform joint training.

\begin{figure}[h]
    \centering
    \includegraphics[width=\textwidth]{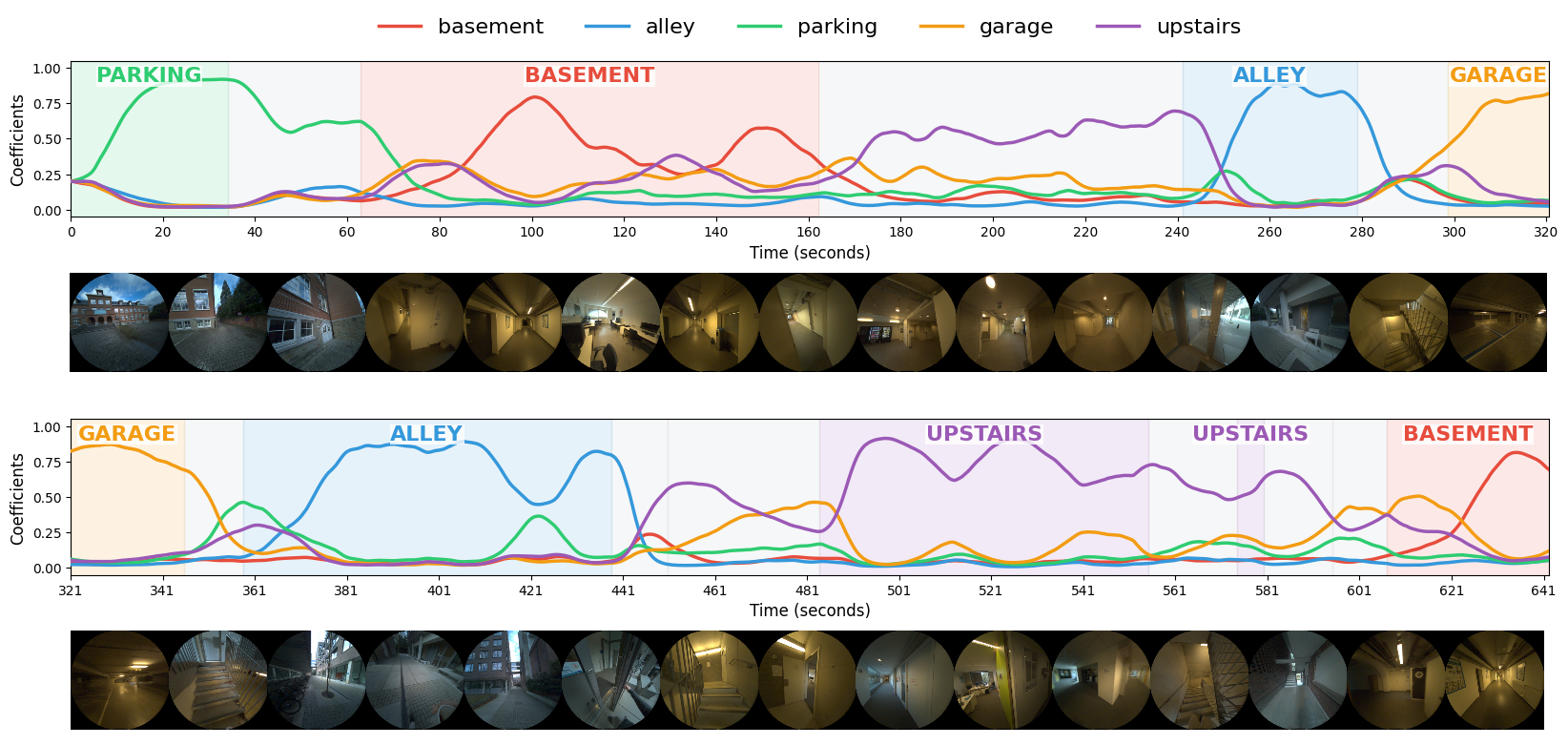}
    \caption{Evolution of the TTT coefficients in the \textbf{Domains-Campus} test stream. Each shaded region represents a domain that was trained on, with its corresponding LoRA coefficient plotted in the same color.}
    \label{fig:coeffs}
\end{figure}

\cref{fig:coeffs} visualizes the behavior of these TTT coefficients $c_k$ over time in the Domains-Campus test stream. We can clearly see that the coefficient corresponding to each domain gains importance when entering its domain. There is some confusion in the 'basement' domain, where at some point the 'upstairs' and 'garage' domain coefficients also contribute significantly. This happens due to some visual similarities between these indoor domains but ends up not hurting performance.

\subsection{Ablation}
\begin{figure}[h]
    \centering
    \includegraphics[width=\textwidth]{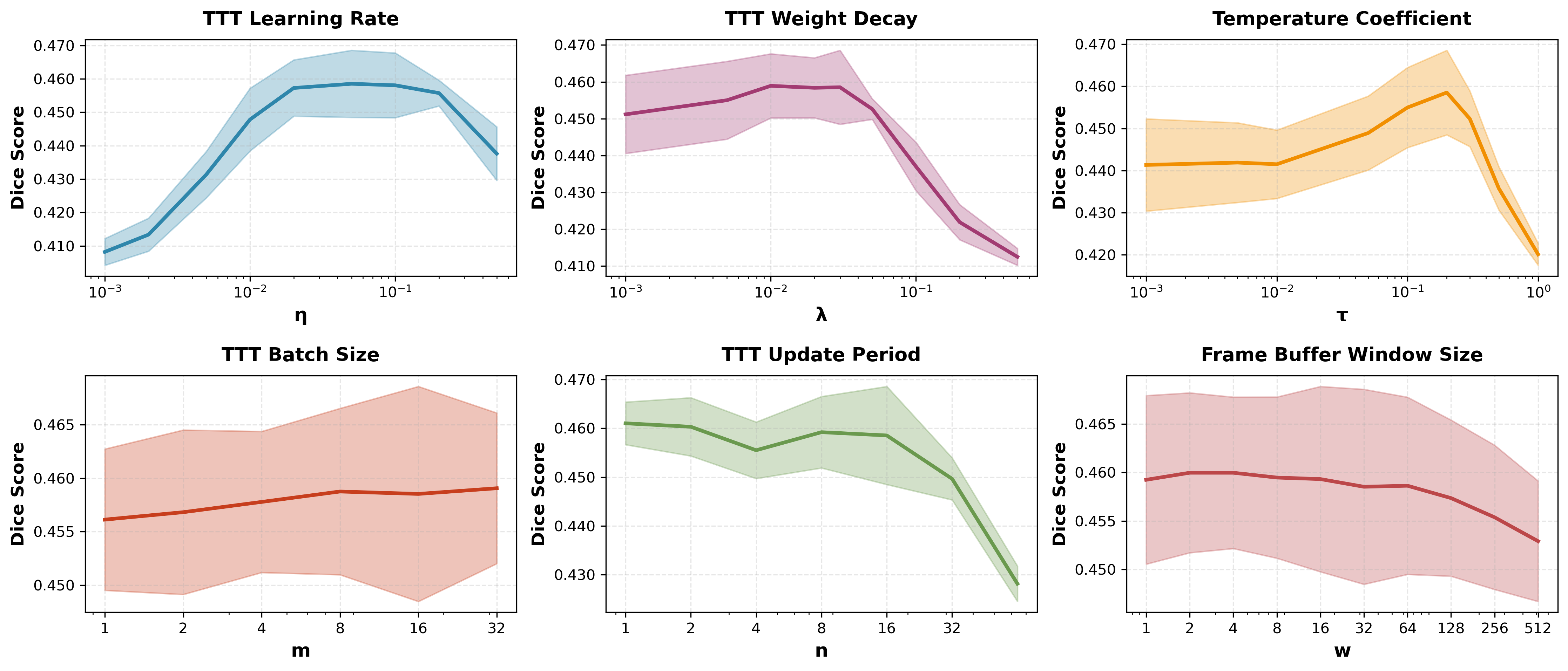}
    \caption{Effect of key hyperparameters of the test-time training procedure. Each plot varies a single hyperparameter while keeping the others fixed to their default value ($\eta=0.05$, $\lambda=0.03$, $\tau=0.2$, $m=16$, $n=16$, $w=32$).
    We plot the mean performance over 3 random training seeds and 3 inference seeds per training seed, for a total of 9 evaluations per config. The shaded band shows $\pm 1$ standard deviation. }
    \label{fig:ablation}
\end{figure}

We perform a series of ablations on the \textbf{Domains-Campus} test stream to analyze the effect of some key hyperparameters of our test-time training procedure. These are visualized in \cref{fig:ablation}. For the learning rate, we observe stable and strong performance in the range $\eta \in [0.02, 0.2]$. Smaller values lead to overly slow adaptation to domain shifts, while larger values make the adaptation unstable. A moderate amount of weight decay also seems to help slightly, with values of $\lambda$ between $0.005$ and $0.05$ yielding the best results. In this regime, the coefficients $c_k$ remain smaller, and thus need less adaptation steps to switch between domains. This synergizes with the coefficient softmax temperature, which then re-amplifies the most relevant coefficient. Excessive weight decay, however, suppresses the coefficients too strongly and makes adaptation too difficult. A softmax temperature in the range $\tau \in [0.1,0.3]$ substantially improves performance by increasing the relative importance of the largest coefficient while still allowing secondary LoRAs to also contribute, potentially allowing some knowledge transfer across similar domains. Very low temperatures make the coefficients overly confident, which reduces performance, whereas high temperatures approach uniform averaging and degrade performance toward the naive LoRA merging baseline. 
Next, the test-time training seems to work quite robustly even with small batch sizes. Larger batch sizes seem to improve performance only slightly. For the update period, one update every $n=16$ frames seems to provide the best trade-off between adaptation speed and compute overhead. Increasing the update period to $32$ or more frames slows down adaptation too much and decreases performance, while the gains from updating every one or two frames are too small to be conclusive, and not worth the significantly increased computational overhead.
Finally, the frame buffer window size does not need to be very large. In fact, the test-time training surprisingly even already works well with a window size of 1, simply training on the current frame only. Window sizes longer than 32 frames only reduce performance.

\section{Discussion}
\label{sec:discussion}

Our experiments show that the proposed method performs strongly compared to other baselines. We attribute this to leveraging the temporal continuity of streaming test data, allowing the model to exploit locality \cite{wang2025test}. While an expert model fine-tuned to a specific setting may outperform a generalist model trained with global replay in that setting, its performance degrades when the domains shift. Our MAE-based test-time-training effectively detects and compensates for such degradation. Locality is often overlooked in continual learning: most benchmarks assume IID test data, rewarding models that “instantly remember everything”. In reality, data in the real world often arrives as a continuous stream, where temporary forgetting can be beneficial to focus on what matters in the present, as long as the model can later easily recover the lost information. Just as humans might need some time to refocus when switching tasks, models might benefit from the same temporal adaptability. Embracing this continuous nature of the real world could yield a more realistic perspective on continual learning.

This work does have some limitations, however. First of all, it is specifically designed for streaming data, but due to the test-time training, parallelization and high-throughput offline processing is more challenging. Test-time training also adds computational and memory overhead during inference, which grows with the number of LoRAs, potentially limiting scalability when dealing with a large number of domains. 
This could be addressed in future work, for example by coming up with ways to reduce the number of LoRAs by merging some of them when possible or by pruning them when they are no longer useful. Another direction could be to look for more efficient ways to decide when test-time training updates are needed. Our works simply performs a training step every $n$ frames, but perhaps we could do better, and only adapt the model when a domain change is detected, and stop updating when the adaptation has converged.
\section{Conclusion}
\label{sec:conclusion}
In summary, we have presented a domain-incremental learning method designed for continuous, streaming test data, such as video streams. Our approach trains domain-specific LoRAs for each domain and leverages test-time training to dynamically select the most appropriate LoRAs during evaluation. This procedure proves effective in maintaining strong performance while minimizing forgetting, demonstrating the potential of combining continual learning with online adaptation in real-world, temporally continuous scenarios.

\newpage
\section*{Acknowledgments}
This paper has received funding from the Flemish Government under the Methusalem Funding Scheme (grant agreement
n° METH/24/009).
\bibliographystyle{splncs04}
\bibliography{main}
\newpage
\appendix  

\section{Implementation Details}
\subsection{Model architectures}
\label{app:arch}

\subsubsection{Video Action Recognition:}
For our action recognition model, the encoder and MAE decoder $E$ and $D$ are based on the VideoMAE architecture \cite{tong2022videomae}. We use a ViT-Tiny model size, and make a few modifications.

First, we replace the original self-attention–based decoder head $D$ with a cross-attention–based decoder, as in CrossMAE \cite{fu2024rethinking}. This modification makes the decoding of each patch independent of the others, allowing us the option to decode only a subset of masked patches. During our pre-training, we thus reconstruct only 50\% of the image patches, randomly selected, instead of the entire image. This significantly reduces memory requirements and speeds up pre-training, allowing us to pre-train the VideoMAE on a single Nvidia L40s GPU.

Secondly, to handle streaming video efficiently, we replace all attention operations with a windowed, causal variant, allowing each patch to attend only to patches from past frames within a sliding window of size $w$. We also replace the positional embedding for the time-axis with RoPE \cite{su2024roformer}. Together, these modifications enable efficient inference on a video stream, processing each incoming frame sequentially using a KV-cache \cite{pope2023efficiently}.

For the main action classification task, we add an additional head $M$ after the encoder, which consists of four attentive classifier blocks, inspired by \cite{chen2024context, bardes2024revisiting}. Each layer consists of cross-attention module followed by an MLP. It takes a learnable classifier token as the query, and uses a learnable weighted sum of output patches from different layers from the encoder as keys and values for cross-attention.

Since the model is trained for action recognition rather than temporal segmentation, it lacks inherent mechanisms to detect action boundaries at test time. Therefore, we assume the start and end times of each action are given. Both at training- and test-time, we only use the final prediction $y_t$ at the last frame of an action clip to determine its class, and the intermediate predictions are ignored. At test-time, we also inform the model a new action has started by clearing its KV-Cache at the beginning of a new clip. 

\begin{figure}[h]
    \centering
    \includegraphics[width=\textwidth]{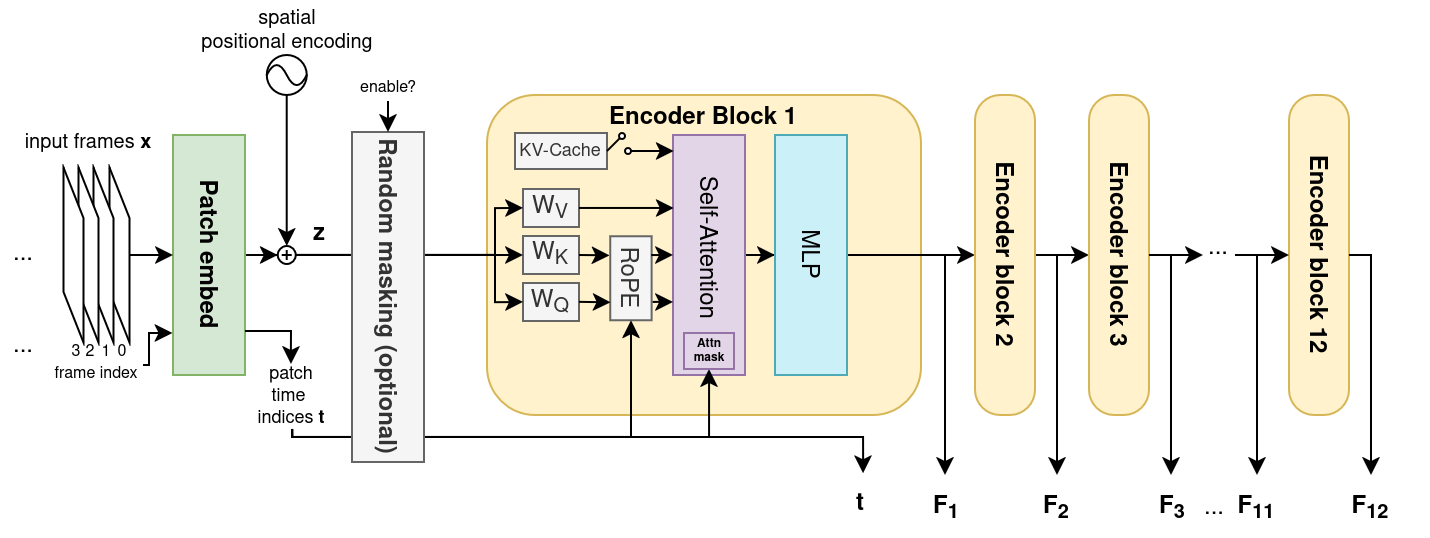}
    \caption{The modified VideoMAE-based encoder architecture. The encoder consists of 12 ViT blocks, where we use RoPE positional encoding for the time axis. The encoder outputs the features $F_i$ of every block $i = 0,1,...,12$. We use an attention mask for causal windowed attention, and KV-Caching is possible for streaming inference}
    \label{fig:enc}
\end{figure}

\begin{figure}[h]
    \centering
    \includegraphics[width=\textwidth]{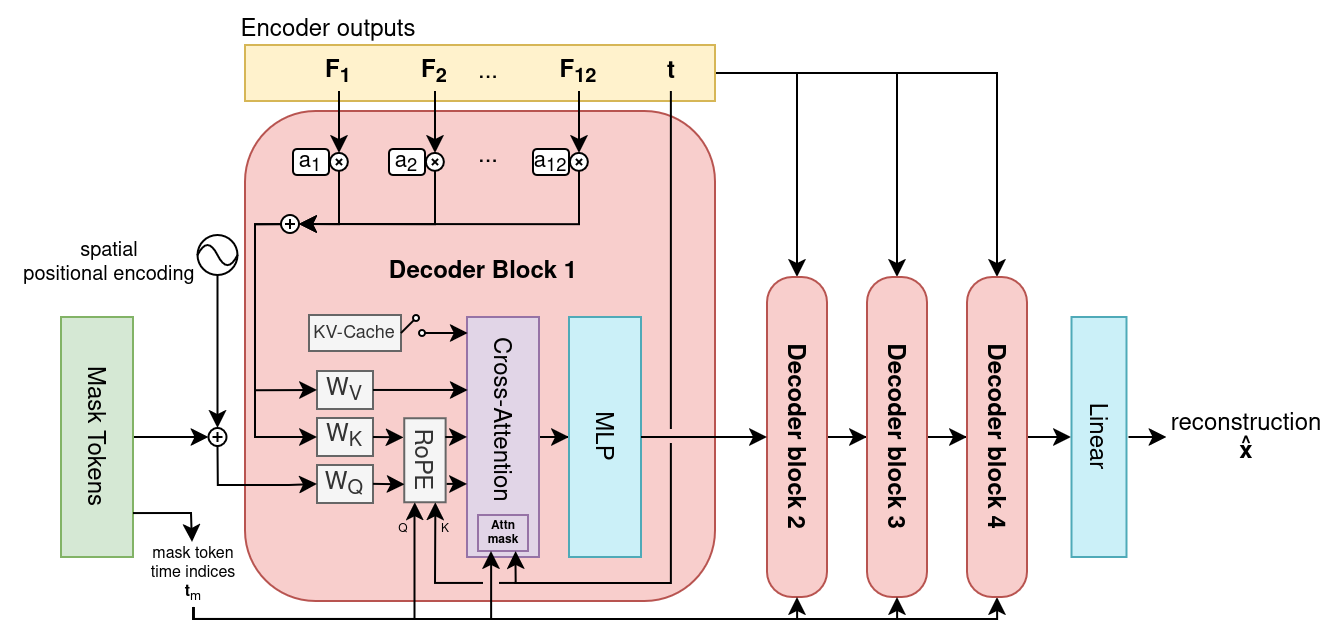}
    \caption{The modified cross-attention decoder architecture, inspired by \cite{fu2024rethinking}. We compute a reconstruction for each of the given mask tokens, which serve as queries for attention. Keys and values are computed from linear combinations of the encoder output features. As there is no self-attention between the mask tokens, each one is reconstructed independently, and we could decide to only reconstruct a subset of the patches instead of the full image. Once again RoPE is used for time position embedding, and we use windowed causal attention with KV-Caching during streaming inference.}
    \label{fig:dec}
\end{figure}

\begin{figure}[h]
    \centering
    \includegraphics[width=\textwidth]{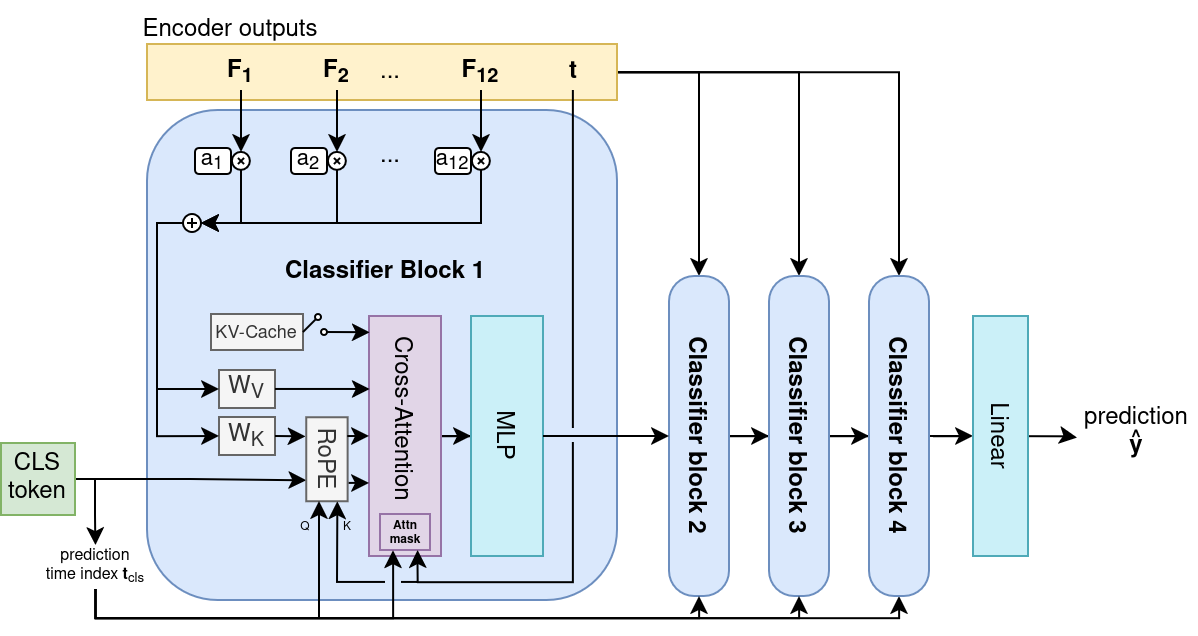}
    \caption{The classifier head for action classification. The architecture is quite similar to the MAE decoder in \cref{fig:dec}, but uses a single classifier token per prediction as query. 
    }
    \label{fig:cls}
\end{figure}

\begin{figure}[h]
    \centering
    \includegraphics[width=\textwidth]{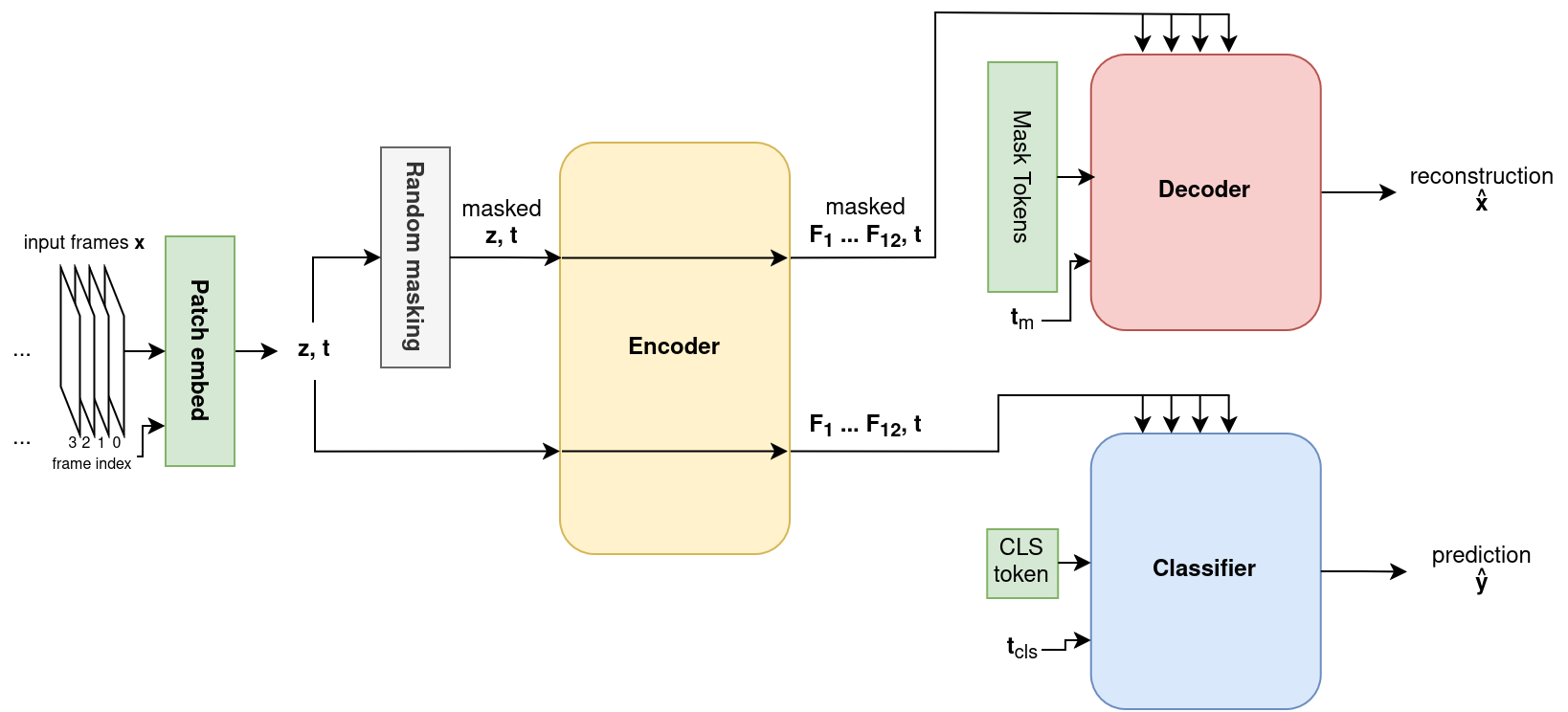}
    \caption{Overview of the entire action classification architecture, combining the encoder, decoder and classifier.}
    \label{fig:actclsarch}
\end{figure}

A full diagram of the model is illustrated in \cref{fig:enc,fig:dec,fig:cls,fig:actclsarch}


\clearpage

\subsubsection{Semantic Segmentation:}
\begin{figure}[h]
    \centering
    \includegraphics[width=\textwidth]{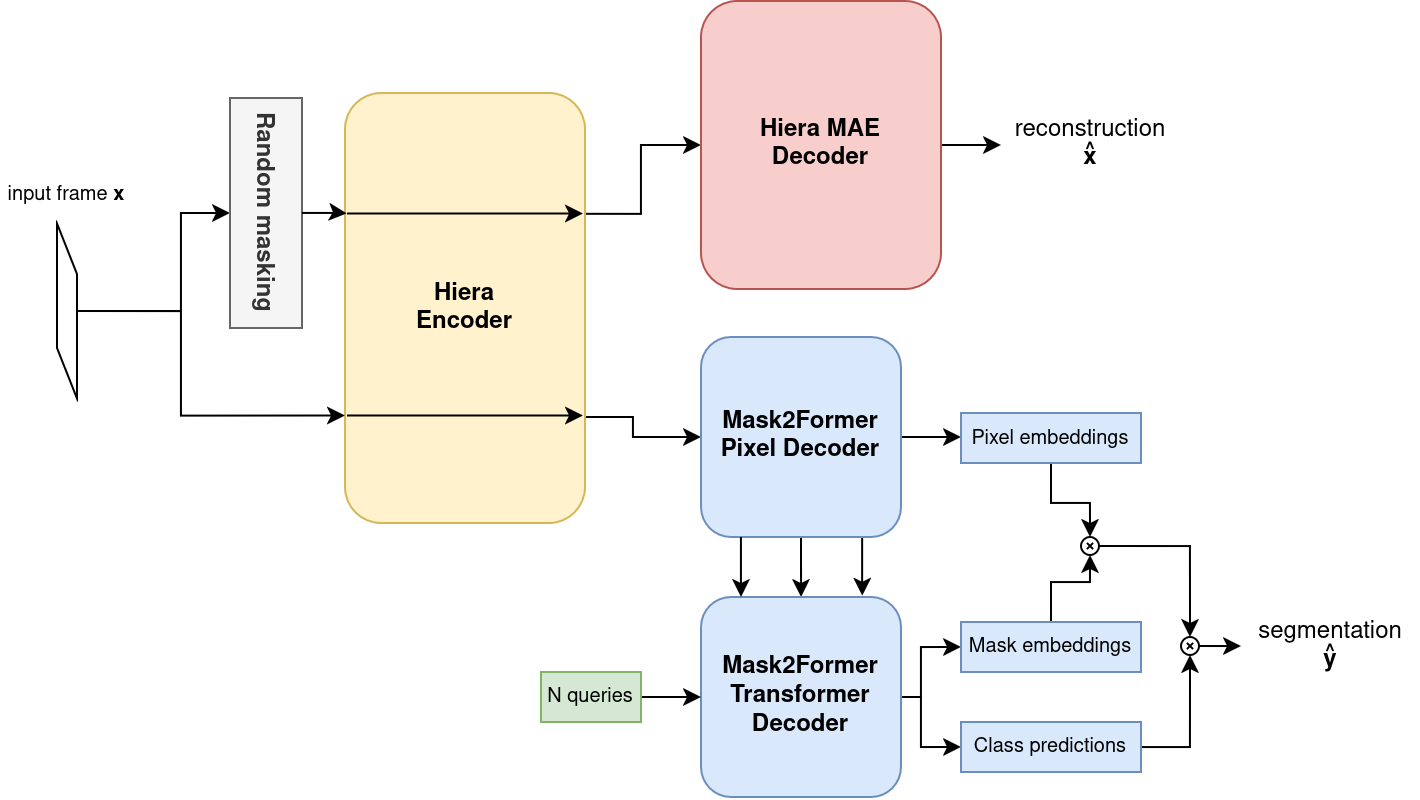}
    \caption{Overview of the entire semantic segmentation architecture. The original Mask2Former Swin backbone encoder is replaced by a Hiera encoder, and we use its corresponding decoder for MAE.}
    \label{fig:m2farch}
\end{figure}

The semantic segmentation model is based on the Mask2Former \cite{Cheng2021MaskedattentionMT} architecture. Its backbone encoder is used as the shared encoder $E$, and the combination of its pixel decoder and transformer decoder for the mask and class predictions act as the main task head $M$. We use the Tiny-size model and we initialize it from a pre-trained checkpoint. The original model uses a Swin-Tiny backbone \cite{liu2021swin}, which is not very well suited for MAE, as we cannot easily "leave out" the masked tokens from the encoder. Prior work \cite{gandelsman2022test, wang2025test} simply blacked out the contents of the masked patches, instead of leaving them out of the input entirely, which works but is computationally wasteful, as this ability to leave out the masked tokens from the encoder is one of the big advantages of the MAE architecture \cite{he2022masked}. To improve this, we replace the backbone with a pre-trained Hiera-Tiny MAE \cite{ryali2023hiera}, which is also a hierarchical ViT like Swin and is straightforward to slot into the original Mask2Former model without any further modifications. We also keep the decoder of this Hiera model's MAE pre-training, and use it as our TTT decoder $D$. A full diagram of the model is illustrated in \cref{fig:m2farch}

An additional implementation detail for training and test-time training on the Domains-Campus involves output masking. The videos in this dataset are circle-shaped (see \cref{fig:tasks}), and are blacked out outside of the circular image. To avoid this influencing the training loss, we use this prior knowledge and perform output masking. Concretely, we completely erase the output segmentation masks in the empty edges of the frame, making it always 'perfectly accurate' (empty) in this region. Additionally, we also only include the patches that are fully inside the center circle in the MAE loss for both offline training and during test-time training. This is so that the MAE only focuses on learning what's actually inside the frame, instead of optimizing the relatively simpler but less interesting task of just making the circular black edge of the frame more accurate.

\subsection{KV-Caching}
\label{app:kv}
As mentioned earlier, the action recognition model uses temporal, causal, windowed attention. This means patches from one frame can only attend to the current or past frames, within a certain time window $w_{KV}$. We do this to allow efficient streaming inference: when we process a video frame by frame, the patches in the past frames do not depend on future frames, so they do not need to be recomputed. We can instead cache \cite{pope2023efficiently} the keys and values of each attention module from the last $w_{KV}-1$ frames. Then each attention module concatenates these cached keys and values to the current input patches' $\mathbf{Z}_{in,t}$ keys and values $\mathbf{K}_t$, $\mathbf{V}_t$ before computing the attention to get the output $\mathbf{Z}_{out,t}$. Our model uses a time window of $w_{KV,ED}=16$ inside the encoder and MAE decoder and $w_{KV,M}=64$ for the classifier.

\begin{align}
\label{eq:kvc}
    \mathbf{Q}_t &= \mathbf{Z}_{in,t} W_K \\
    \mathbf{K}_t &= \mathbf{Z}_{in,t} W_K \\
    \mathbf{V}_t &= \mathbf{Z}_{in,t} W_V \\
    \mathbf{\tilde{\mathbf{K}}}_t &= [\mathbf{K}_{t-w_{KV}+1}, ..., \mathbf{K}_t] \\
    \mathbf{\tilde{\mathbf{V}}}_t &= [\mathbf{V}_{t-w_{KV}+1}, ..., \mathbf{V}_t] \\
    \mathbf{Z}_{out,t} &= Attention(\mathbf{Q}_t, \mathbf{\tilde{\mathbf{K}}}_t, \mathbf{\tilde{\mathbf{V}}}_t)
\end{align}

\begin{figure}[h]
    \centering
    \includegraphics[width=0.5\textwidth]{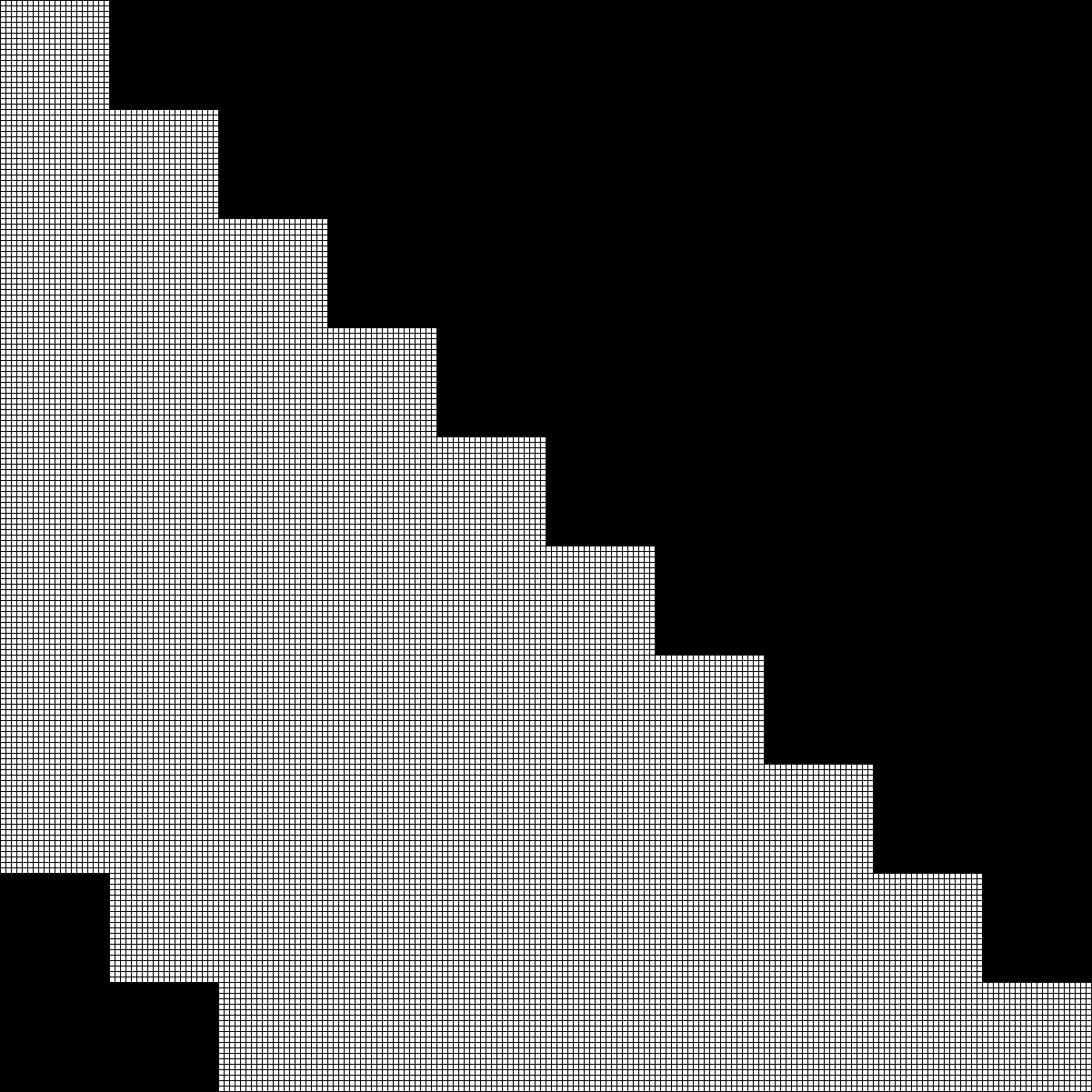}
    \caption{Example causal sliding window attention mask which simulates the KV-cache at training time.}
    \label{fig:attmask}
\end{figure}

At training time, when processing all frames in a batched manner, this is equivalent to applying an attention mask shaped as illustrated in \cref{fig:attmask}. This makes the attention causal and windowed. It blocks tokens from attending to future patches, and also blocks them from attending to tokens too far in the past. Its "blocky" structure stems from the fact that in ViTs, we have several patch tokens that correspond to the same time step.

The temporal attention and KV-caching and requires us to slightly change \cref{alg:online_lora_update}. An updated version is shown in \cref{alg:kv}, where with the expressions in \cref{eq:kvnotation1,eq:kvnotation2} we express using the branch $F$ with a given KV-cache, returning the prediction and the new keys and values of the current timestep $t$.

\begin{align}
\label{eq:kvnotation1}
\mathbf{y}, \mathbf{K}_{E,t} \mathbf{V}_{E,t},\mathbf{K}_{M,t} \mathbf{V}_{M,t} &= F_\text{main}(\mathbf{x}, \mathbf{K}_E, \mathbf{V}_E,\mathbf{K}_M, \mathbf{V}_M) \\
\label{eq:kvnotation2}
\mathbf{\hat{x}}, \mathbf{K}_{E,t} \mathbf{V}_{E,t},\mathbf{K}_{D,t} \mathbf{V}_{D,t} &= F_\text{MAE}(\mathbf{x}, \mathbf{K}_E, \mathbf{V}_E,\mathbf{K}_D, \mathbf{V}_D)
\end{align}

\begin{algorithm}
\caption{Online Adaptation with Test-Time training: KV-Cache version}
\label{alg:kv}
$\mathbf{c} \gets [1/N, 1/N, ..., 1/N]$ \tcp{initialize model LoRA coeffs to average}
$\mathbf{K}_{E,\text{MAE}} \gets []$; \tcp{initialize empty KV-caches}
$\mathbf{V}_{E,\text{MAE}} \gets []$;\\
$\mathbf{K}_{E,\text{main}} \gets []$;\\
$\mathbf{V}_{E,\text{main}} \gets []$;\\
$\mathbf{K}_{D} \gets []$;\\
$\mathbf{V}_{D} \gets []$;\\
$\mathbf{K}_{M} \gets []$;\\
$\mathbf{V}_{M} \gets []$;\\
$\text{optimizer.params} \gets \mathbf{c}$ \tcp{only adapt coeffs, rest of model is frozen}
\For{$t = 1$ \KwTo $T$}{
    $x_t \gets $ Next frame from input stream;\\
    \If{$|B|=w$}{
        Remove $x_{t-w}$ from $B$ \tcp{sliding window buffer is full}
    } 
    Save $x_t$ to $B$;\\
    set $F_{\text{MAE}}$ coeffs $\gets \mathbf{c}$\;

    \If{$t \bmod n = 0$}{ \tcp{perform TTT update on the last $w$ frames} 
        $\mathbf{x_\text{recent}} \gets B$\;
    
        $\mathbf{x_\text{mask}} \gets $ Randomly mask a fraction $r_{mask}$ of patches in $\mathbf{x_\text{recent}}$\;
    
        $\mathbf{\hat{x}}_\text{dec}, \mathbf{K}_{E,\text{MAE},t} \mathbf{V}_{E,\text{MAE},t},\mathbf{K}_{D,t} \mathbf{V}_{D,t} \gets F_\text{MAE}(\mathbf{x}_\text{recent}, \mathbf{K}_{E,\text{MAE}}, \mathbf{V}_{E,\text{MAE}},\mathbf{K}_D, \mathbf{V}_D)$\;
        $L \gets \mathcal{L}_{\text{MAE}}(\mathbf{\hat{x}_\text{dec}}, \mathbf{x_\text{recent}})$ \;
        $\mathbf{c} \gets \text{optimizer.step}(\mathbf{c}, \nabla_\mathbf{c} L)$         \tcp{update model LoRA coeffs}
    }
    \ElseIf{$0 < t \bmod n \le n-w$ }{
        \tcp{These frames would be skipped by TTT, evaluate them anyways to update KV-cache}
         $\_, \mathbf{K}_{E,\text{MAE},t} \mathbf{V}_{E,\text{MAE},t},\mathbf{K}_{D,t} \mathbf{V}_{D,t} \gets F_\text{MAE}(x_t, \mathbf{K}_{E,\text{MAE}}, \mathbf{V}_{E,\text{MAE}},\mathbf{K}_D, \mathbf{V}_D)$\;
    }
    \Else{
        \tcp{The last $m$ frames of the update period will be processed by the MAE later, in the next the update step.}
        $\mathbf{K}_{E,\text{MAE},t} = []$; \tcp{will be computed later}
        $\mathbf{V}_{E,\text{MAE},t} = []$;
    }
    \tcp{make prediction y}
    set $F_{\text{main}}$ coeffs $\gets softmax(\mathbf{c}/\tau)$\;
    $\hat{y}_t, \mathbf{K}_{E,\text{main},t} \mathbf{V}_{E,\text{main},t},\mathbf{K}_{M,t} \mathbf{V}_{M,t}
 \gets F_{\text{main}}(x_t, \mathbf{K}_{E,\text{main}}, \mathbf{V}_{E,\text{main}},\mathbf{K}_M, \mathbf{V}_M)$

    \tcp{manage KV-Cache}
    remove keys \& values before $t-w_{DE}+2$ from $\mathbf{K}_{E,\text{MAE}},\mathbf{V}_{E,\text{MAE}}$;\\
    add $\mathbf{K}_{E,\text{MAE},t},\mathbf{V}_{E,\text{MAE},t}$ to $\mathbf{K}_{E,\text{MAE}},\mathbf{V}_{E,\text{MAE}}$;\\
    remove keys \& values before $t-w_{DE}+2$ from $\mathbf{K}_{E,\text{main}},\mathbf{V}_{E,\text{main}}$;\\
    add $\mathbf{K}_{E,\text{MAE},t},\mathbf{V}_{E,\text{MAE},t}$ to $\mathbf{K}_{E,\text{main}},\mathbf{V}_{E,\text{main}}$;\\
    remove keys \& values before $t-w_{DE}+2$ from $\mathbf{K}_{D},\mathbf{V}_{D}$;\\
    add $\mathbf{K}_{D,t},\mathbf{V}_{D,t}$ to $\mathbf{K}_{D},\mathbf{V}_{D}$;\\
    remove keys \& values before $t-w_{M}+2$ from $\mathbf{K}_{M},\mathbf{V}_{M}$;\\
    add $\mathbf{K}_{M,t},\mathbf{V}_{M,t}$ to $\mathbf{K}_{M},\mathbf{V}_{M}$;\\
}
\end{algorithm}

At inference, we maintain separate KV-caches for the main task branch $F_{\text{main}}$ and for the MAE branch $F_{\text{MAE}}$. To keep the KV-cache of the MAE branch up to date, we now also evaluate the MAE branch when no TTT backpropagation step is required. To correctly use the temporal attention, we also do not randomly sample frames. Instead, we take the entire frame buffer in the correct order, and use this as a video clip input for the VideoMAE. This means that the frame buffer window and TTT batch size $w=m$ are the same. For simplicity, we also constrain this window size to the update period, $w \le n$, such that every frame is processed exactly once and in the correct order, which would otherwise complicate the KV-cache management further by creating duplicate entries in the cache.

\subsection{Training recipes}
\label{app:train}
\subsubsection{Action Classification:}
For the \textbf{MAE pre-training} of $E,D$, we train on the Kinetics400 \cite{kay2017kinetics} dataset using a recipe similar to \cite{tong2022videomae}, summarized in \cref{tab:maeprercp}. We sample dense segments of 32 frames with a stride of 2 from the videos, from which we mask 90\% of the patches, which we then reconstruct.
Due to the modified decoder, we can use a lower reconstruction ratio, which we set to 50\%, leading to lower VRAM requirements, allowing us to train the model on a single Nvidia L40s GPU in roughly 30 hours.

\begin{table}
    \centering
    \caption{\textbf{Kinetics400 pre-training}
    }
    \footnotesize
    \begin{tabular}{c|c}
        config & value \\
    \midrule
        epochs & 800 \\
        optimizer & AdamW \\
        lr & 1.5e-4 \\
        weight decay & 5e-2 \\
        betas & (0.9, 0.95) \\
        batch size & 256 \\
        lr schedule & cosine decay \\
        warmup epochs & 40 \\
        masking ratio & 0.95 \\
        reconstruction ratio & 0.5 \\
        augmentations & \makecell{flip, multiscale crop} \\
    \end{tabular}
    \label{tab:maeprercp}
\end{table}

Next, \textbf{finetuning the initial task} also uses a recipe similar to \cite{tong2022videomae}, summarized in \cref{tab:actftrcp}. One thing we must note is our frame sampling strategy during training. First of all, when reading frames from the video files, we use a stride of 2 with temporal jitter of $\pm 1$. Since the videos are stored at 20 FPS, this means the resulting action clip inputs for the model run at 10 FPS.

\begin{figure}[h]
    \centering
    \includegraphics[width=\textwidth]{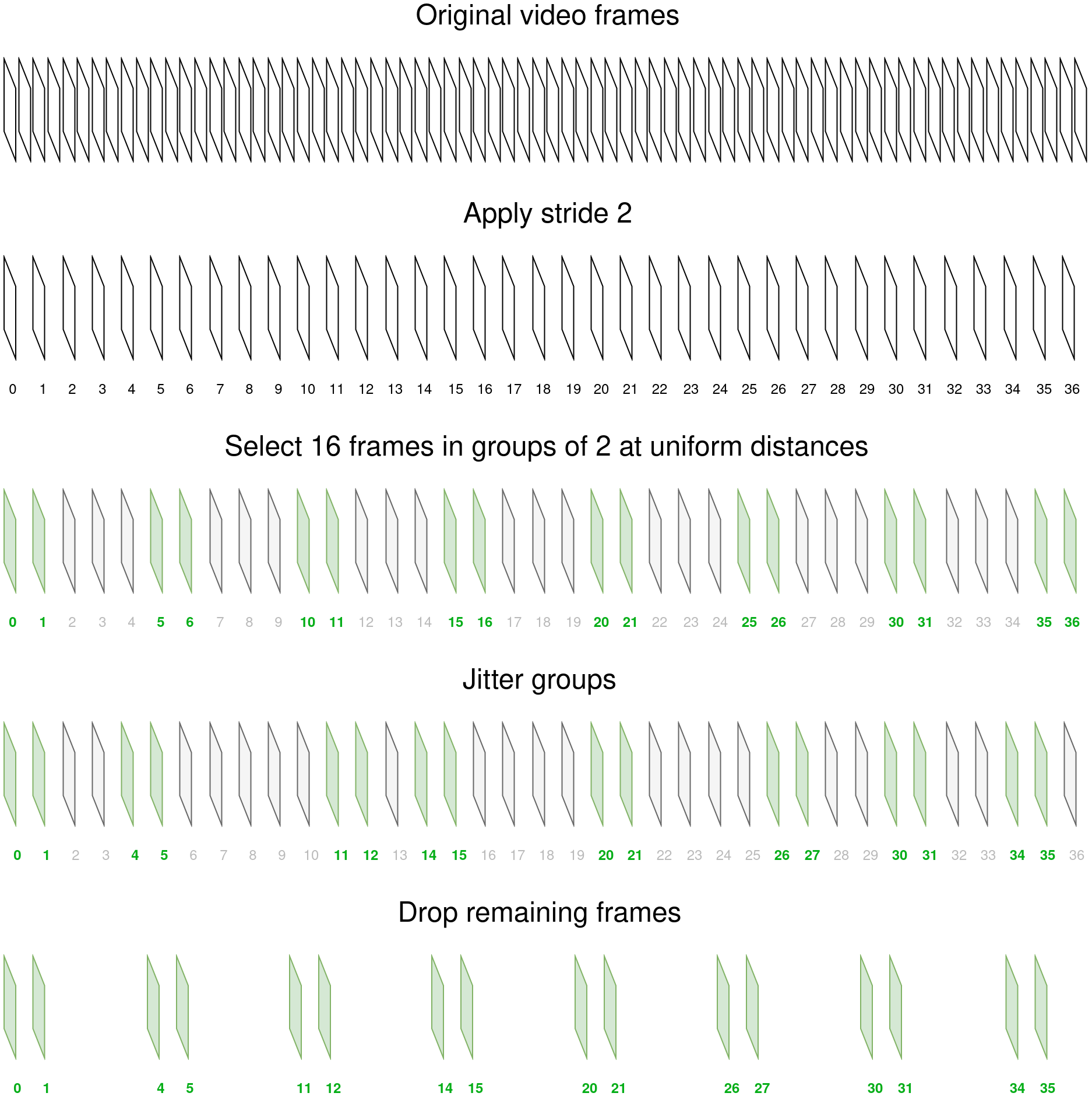}
    \caption{Sparse frame sampling strategy used to speed up training the first 50 epochs of $T_0$.}
    \label{fig:sampl}
\end{figure}

To further speed up training, we train the first 50 epochs on \emph{sparsely} sampled clips, using only 16 frames per action clip, while the other frames are dropped. This dropping of frames is done in a very specific way, keeping into account the architecture of the MAE model. Concretely, the VideoMAE model's patch embedding first tokenizes the video using a 3D convolution with a kernel depth of 2 on the time axis, meaning each resulting spatiotemporal patch contains information of 2 consecutive frames (at 10 FPS). To not interfere with this behaviour, we thus make sure our sparse sampling selects groups of 2 consecutive frames as well, instead of standard uniform sampling. These groups are first sampled such that the distance between the groups is uniform, and are then slightly jittered. We also keep track of the frame indices to correctly apply the same RoPE positional embeddings to the remaining frames as if no frames were dropped. The procedure is illustrated in \cref{fig:sampl}.

After these first 50 epochs, we restart the training schedule and finetune the model for one more epoch on \emph{full-length clips}. Due to the fact that clips can be variable length and are thus difficult to batch, we use a real batch size of 1, but 'simulate' it to be larger by using a bigger gradient accumulation. 

A second thing to note is the use of drop path \cite{huang2016deep}. This is a standard technique to reduce overfitting when finetuning ViTs. We only apply drop path in the main task branch. In the MAE branch, drop path is disabled.
We also adopt layer-wise learning rate decay \cite{bao2021beit}, once again as is standard for finetuning MAEs \cite{he2022masked,tong2022videomae}. This is only applied to the encoder $E$. The decoder $D$ and main task head $M$ are all trained at the base learning rate.

The MAE reconstruction ratio was set to the highest amount that fit in the VRAM of the GPU used for finetuning, which was a single Nvidia RTX 3090. The process takes roughly 8 hours.

Finally, for continual learning, we use roughly the same recipe, with some slight changes. These changes are summarized in \cref{tab:actlorarcp}. Specifically, we add LoRA dropout, randomly leaving out a percentage of the LoRAs, and we train for a shorter training schedule with a lower effective batch size (lower gradient accumulation). Parameters that are not mentioned in this table remain unchanged from \cref{tab:actftrcp}.
We only train on full length action clips. We do not perform sparse frame sampling.
We again use a single Nvidia RTX 3090 GPU for training. Training all 8 tasks takes roughly 10 hours.

For the training of the full finetuning baselines (Naive FT, Replay, DRIFT), we also use this same recipe (ignoring the LoRA dropout which is irrelevant for these baselines), however, we reduce the learning rate to 2.5e-5, as we observed empirically that a higher learning was less stable and caused more extreme forgetting. For baselines involving a replay buffer (Replay, DRIFT), we use a buffer size of 300.

\begin{table}
    \centering
    \caption{\textbf{NTU-RGB+D120 $\mathbf{T_0}$ finetuning}
    }
    \footnotesize
    \begin{tabular}{c|c|c}
        config & value (sparse stage) & value (full-length stage) \\
    \midrule
        epochs & 50 & 1 \\
        optimizer & AdamW & AdamW \\
        lr & 5e-4 & 5e-4 \\
        weight decay & 1e-2 & 1e-2 \\
        betas & (0.9, 0.999) & (0.9, 0.999) \\
        batch size & 16 & 1 \\
        gradient accumulation & 8 & 128 \\
        layer-wise lr decay ($E$) \cite{bao2021beit} & 0.7 & 0.7 \\
        label smoothing \cite{szegedy2016rethinking} & 0.1 & 0.1 \\
        drop path ($F_\text{main}$)\cite{huang2016deep}  & 0.1 & 0.1 \\
        lr schedule & cosine decay & cosine decay \\
        warmup epochs & 5 & 0.1 \\
        masking ratio & 0.9 & 0.9 \\
        reconstruction ratio & 0.4 & 0.9 \\
        gradient clipping & 10 & 10 \\
        augmentations & \makecell{flip, multiscale crop,\\temporal jitter} & \makecell{flip, multiscale crop,\\temporal jitter}\\
    \end{tabular}
    \label{tab:actftrcp}
\end{table}

\begin{table}
    \centering
    \caption{\textbf{NTU-RGB+D120 LoRA finetuning}
    }
    \footnotesize
    \begin{tabular}{c|c}
        config & value \\
    \midrule
        epochs & 15 \\
        lr & 2.5e-4 \\
        batch size & 1 \\
        gradient accumulation & 64 \\
        warmup epochs & 2 \\
        masking ratio & 0.9 \\
        reconstruction ratio & 0.9 \\
        lora dropout & 0.15 \\
        gradient clipping & 10 \\
        augmentations & \makecell{multiscale crop,\\temporal jitter}
    \end{tabular}
    \label{tab:actlorarcp}
\end{table}

\subsubsection{Semantic Segmentation}
As mentioned before, we initialize the semantic segmentation model based on pretrained checkpoints, so we do not need to pretrain the MAE backbone from scratch. We immediately finetune it on $T_0$ with the recipe summarized in \cref{tab:segftrcp}. This training was done on a single Nvidia A100 GPU, and takes roughly 50 hours.

Next, we train the LoRAs of the 5 domains from Domains-Campus. Slight changes are made to the training recipe for this continual training, which mainly involves using a higher learning rate and a lower batch size, as summarized in \cref{tab:seglorarcp}. LoRA training is done using an Nvidia RTX 3090 GPU, and takes roughly 5 hours.

Once again, the full finetuning baselines use this modified recipe as well, but keep the learning rate lower (1e-5 for $E$ and 1e-4 for $D,M$), since we observed empirically that the higher learning rate caused more extreme forgetting. The replay-based baselines use a buffer size of 200. 

\begin{table}
    \centering
    \caption{\textbf{Segmentation $\mathbf{T_0}$ (Places365) finetuning}
    }
    \footnotesize
    \begin{tabular}{c|c}
        config & value \\
    \midrule
        epochs & 15 \\
        optimizer & AdamW \\
        lr ($E$) & 4e-5 \\
        lr ($D,M$) & 4e-4 \\
        weight decay & 5e-2 \\
        betas & (0.9, 0.999) \\
        batch size & 64 \\
        lr schedule & cosine decay \\
        warmup epochs & 1.5 \\
        masking ratio & 0.6 \\
        gradient clipping & 0.1 \\
        augmentations & flip, random resized crop \\
    \end{tabular}
    \label{tab:segftrcp}
\end{table}

\begin{table}
    \centering
    \caption{\textbf{Segmentation (Domains-Campus) LoRA finetuning}
    }
    \footnotesize
    \begin{tabular}{c|c}
        config & value \\
    \midrule
        epochs & 20 \\
        lr ($E$) & 1e-4 \\
        lr ($D,M$) & 1e-3 \\
        batch size & 16 \\
        warmup epochs & 4 \\
        masking ratio & 0.6 \\
        lora dropout & 0.15 \\
        gradient clipping & 0.1 \\
        augmentations & flip, random resized crop \\
    \end{tabular}
    \label{tab:seglorarcp}
\end{table}

\newpage
\section{Dataset details}
\subsection{Action Classification}
\label{app:ntu}
The \textbf{NTU-RGB+D120} dataset \cite{shahroudy2016ntu, liu2019ntu} consists of short recordings of 120 different classes of human actions, performed by a group of participants in several different environments and from a variety of different camera viewpoints. In the first version of this dataset, consisting of the first 60 action classes, there is less variation in the different environments, which makes it less ideal for domain incremental learning experiments. Its extension, consisting of classes 61-120, which we use for our experiments, is better suited, as there is more variety in the locations where actions were recorded.

For the action recognition experiment, we define a domain as the combination of a specific recording location and camera angle. The domain does \textbf{not} depend on the participant performing the actions: multiple participants perform actions within a given domain, and some participants are recorded in multiple domains.

The initial task $T_0$ is a more varied combination of all camera angles from setups \emph{S018, S019, S026, S027} and \emph{S031}. These are recorded in the same room. 

The next two tasks, $T_1$ and $T_2$ are each a single camera angle in setups \emph{S029C003} and \emph{S30C002} respectively, which are recorded in the same room as $T_0$ but in a different direction and from a very different camera angle.

Next, $T_3$ uses data recorded in a different location from camera angles \emph{S021C002} and \emph{S022C002}, which are combined as they are very similar, and on their own they otherwise have significantly less samples than the other tasks.

$T_4$ similarly also combines data from \emph{S021C003} and \emph{S022C003}, recorded in the same location as $T_4$ but from the opposite direction.

$T_5$ and $T_6$ are recorded in the next setup, \emph{S023C002} and \emph{S23C003}, both in the same location but again from opposite directions.

Finally, the last two tasks again both combine the data of 2 very similar setups, with \emph{S024C002} and \emph{S025C002} making up $T_7$ and \emph{S024C003} and \emph{S025C003} for $T_8$. The remaining camera setups from the original dataset that were not mentioned are not used in any experiments.

We split the data into a train and a test set based on the ID of the participant performing the actions. Thus, any participant that occurs in the training split does not occur in the test set and vice versa.
The IDs of the participants in the train set are [8,42,43,45,46,47,49,50,52,53,54,55,56,57,58,59,61,62,63,65,67,70,72, 73,74,75,76,77,78,80,81,82,83,84,85,86,87,88,89,90,91,92,93,94,95,97,98,99,100,\\ 101,103,104,105,106].
For the test set we use IDs [6,11,41,44,48,51,60,64,66,68,69, 71,79,96,102]. The remaining participants that are in neither of these lists do not occur in the CL task sequence. This train/test split is different from the original cross-participant evaluation splits proposed by the dataset creators. This is to have a better balance of train versus test data in each individual task.

Before use, the videos were downsampled and stored at a resolution of $640 \times 320$ pixels with 20 frames per second.

\subsection{Semantic Segmentation}
\label{app:campus}
For our semantic segmentation experiment, we wish to predict binary segmentation maps of 32 different concepts. The chosen concepts are listed in \cref{tab:segcl}

\begin{table}[h]
\caption{Segmentation concepts}
\label{tab:segcl}
\centering
\begin{tabular}{p{0.05\textwidth} p{0.4\textwidth} p{0.05\textwidth} p{0.5\textwidth}}
\textbf{ID} & \textbf{Class} & \textbf{ID} & \textbf{Class} \\
\toprule
0  & human & 16 & bush \\
1  & wall & 17 & grass \\
2  & ceiling & 18 & path \\
3  & floor & 19 & table, desk \\
4  & building & 20 & chair \\
5  & window & 21 & mug, cup \\
6  & door, doorway & 22 & computer \\
7  & backpack & 23 & paper sheet \\
8  & trashcan & 24 & book \\
9  & sign & 25 & writing board, whiteboard, blackboard \\
10 & phone & 26 & wire, cable \\
11 & poster, picture & 27 & cabinet \\
12 & bottle & 28 & box \\
13 & bike & 29 & pen, pencil \\
14 & car & 30 & stairs, staircase \\
15 & tree & 31 & lighting \\
\end{tabular}
\end{table}

For $T_0$, we train on a subset of the \textbf{Places365} \cite{zhou2017places}. This is a collection of images captured in various types of locations. We specifically select a subset of types of locations that are relevant to the concepts we wish to segment. These locations are listed in \cref{tab:plsubset}

\begin{table}[h]
\caption{Subset of Places365 location types used for $T_0$}
\centering
\scriptsize
\begin{tabular}{llll}
alley & department\_store & hotel\_room & park \\
apartment\_building-outdoor & dining\_hall & house & parking\_garage-indoor \\
archive & dining\_room & inn-outdoor & parking\_lot \\
art\_gallery & doorway-outdoor & japanese\_garden & patio \\
art\_school & dorm\_room & kindergarden\_classroom & physics\_laboratory \\
art\_studio & downtown & kitchen & picnic\_area \\
artists\_loft & driveway & lawn & playroom \\
auditorium & elevator\_lobby & lecture\_room & pond \\
balcony-exterior & embassy & legislative\_chamber & porch \\
balcony-interior & entrance\_hall & library-indoor & pub-indoor \\
banquet\_hall & fastfood\_restaurant & library-outdoor & reception \\
bar & fire\_escape & living\_room & recreation\_room \\
biology\_laboratory & forest-broadleaf & lobby & restaurant \\
bookstore & forest\_path & locker\_room & staircase \\
booth-indoor & forest\_road & mansion & storage\_room \\
bow\_window-indoor & formal\_garden & manufactured\_home & street \\
building\_facade & garage-indoor & mezzanine & supermarket \\
cafeteria & garage-outdoor & motel & sushi\_bar \\
campus & gas\_station & movie\_theater-indoor & television\_room \\
chemistry\_lab & general\_store-outdoor & museum-indoor & television\_studio \\
classroom & hardware\_store & museum-outdoor & train\_interior \\
coffee\_shop & highway & nursery & utility\_room \\
computer\_room & home\_office & nursing\_home & waiting\_room \\
conference\_center & home\_theater & office & wet\_bar \\
conference\_room & hospital & office\_building & yard \\
corridor & hospital\_room & office\_cubicles & youth\_hostel \\
cottage & hotel-outdoor & pantry &  \\
\end{tabular}
\label{tab:plsubset}
\end{table}

We use SAM3 \cite{carion2025sam} to generate a synthetic segmentation map for each of the 32 concepts for every image. Rather than binarizing the outputs, we retain soft labels to preserve the SAM3 model’s confidence information. Specifically, we take SAM3’s mask logits and apply a sigmoid to produce a soft probability for each class at every pixel. To reduce storage, we keep only the three highest probabilities per pixel, as most classes typically have near-zero scores. Images containing very few pixels corresponding to the concepts are discarded. The resulting $T_0$ dataset consists of roughly 540k images and segmentation maps, each stored at a resolution of $288 \times 288$ pixels.

For the CL task sequence $T_1$-$T_5$, we recorded the \textbf{Domains-Campus} dataset, which consists of a set of egocentric videos we recorded inside and around a university building. These videos were recorded and preprocessed using Meta ARIA glasses \cite{engel2023project} and their corresponding software tools. We recorded one training video for each of the 5 domains: 'basement', 'alley', 'parking', 'garage' and 'upstairs'. The length of these videos varies between roughly 2.5 to 5 minutes. We additionally recorded a longer test stream, which is a continuous walk of roughly 11 minutes, passing through each of the 5 domains. The original videos run at a framerate of 30 FPS and were first rectified and rescaled to a resolution of $1408 \times 1408$.

Next, we used SAM3 in video inference mode to generate segmentation maps for our chosen concepts, once again at a resolution of $288 \times 288$. 

We noticed that SAM3 tended to make certain mistakes which we could easily fix manually. Specifically, SAM3 would often predict pixels corresponding to 'floor' to be both 'wall' and 'floor', which should not be possible. Similarily, 'wall' and 'ceiling', and 'wall' and 'door' were often confused. In these cases where 'wall' and one of these other classes were predicted, we erased the wall class. Another case of confusion was the occurrence of a smartphone in the videos, which was often classified as both a 'phone' and 'book'. This was also fixed by erasing the 'book' class when both these two classes were predicted simultaneously.

After extracting the segmentation maps, we blurred sensitive information (e.g. faces, license plates,...) and downscaled the videos to $288 \times 288$.

We split up each training video into chunks of 1 second. Next, we randomly select 20\% of these chunks to use as a validation split during training. We do not train on these validation chunks, and also leave out the chunks immediately before and after each of them. From the remaining chunks, we uniformly sample 4096 frames which are used as training split. For evaluation, we use all frames of the test stream video.

\end{document}